\RequirePackage{amsmath}
\documentclass{llncs}
\usepackage[utf8]{inputenc}
\usepackage{float}
\usepackage{algorithm}
\usepackage{amssymb}
\usepackage{graphicx}
\usepackage{subfigure}
\usepackage[inline]{enumitem}
\usepackage{algpseudocode}
\usepackage{color}
\usepackage{soul}

\title{Modeling Taxi Drivers' Behaviour for the Next Destination Prediction}
\author{Alberto Rossi\inst{1} \and Gianni Barlacchi\inst{2,4} \and Monica Bianchini\inst{3} \and Bruno Lepri\inst{4}}
\institute{University of Florence, Florence, Italy \email{alberto.rossi@unifi.it} \and University of Trento, Trento, Italy \email{gianni.barlacchi@gmail.com} \and University of Siena, Siena, Italy \email{monica@diism.unisi.it} \and Fondazione Bruno Kessler, Trento, Italy \email{lepri@fbk.eu}}
\begin{document}
\maketitle
\begin{abstract}

In this paper, we study how to model taxi drivers' behaviour and geographical information for an interesting and challenging task: the next destination prediction in a taxi journey. Predicting the next location is a well studied problem in human mobility, which finds several applications in real-world scenarios, from optimizing the efficiency of electronic dispatching systems to predicting and reducing the traffic jam. This task is normally modeled as a multiclass classification problem, where the goal is to select, among a set of already known locations, the next taxi destination. We present a Recurrent Neural Network (RNN) approach that models the taxi drivers' behaviour and encodes the semantics of visited locations by using geographical information from Location-Based Social Networks (LBSNs). In particular, RNNs are trained to predict the exact coordinates of the next destination, overcoming the problem of producing, in output, a limited set of locations, seen during the training phase. The proposed approach was tested on the ECML/PKDD Discovery Challenge 2015 dataset --- based on the city of Porto ---, obtaining better results with respect to the competition winner, whilst using less information, and on Manhattan and San Francisco datasets.
\end{abstract}

\section{Introduction}
\label{sec1}
World population is increasingly moving from rural areas to urban centers, making large cities densely populated. Actually, approximately $54\%$ of people worldwide live in cities, especially in metropolis, which offer better working and leisure opportunities. For instance, in urban areas there is greater access to work, a wide variety of options for education and training, ease of transport and abundance of attractive places within a few kilometers. Across huge cities people tend to move more and have to do it faster than in the past. On the other hand, heavy traffic (e.g., traffic jams, severe traffic congestions, etc.) can cause noise and atmospheric pollution (i.e., smog, hydrocarbon concentration, and exhaust gas emissions). Optimizing the public transportation system can therefore help in improving the quality of citizens' lives, both by facilitating their mobility and ensuring their health. The research in this field can also support transport companies to give a better service, in terms of waiting times for the customers, fuel saving, traffic reduction, and ease of mobility. 

In a smart city environment, urban data are captured by sensors, actuators, and mobile devices, and then analyzed based on network infrastructures \cite{ferretti2018weak}. It goes without saying that using such data opens the door to several applications, including forecasting of urban flows in order to improve and integrate the dimensions of the physical, digital and institutional spaces of a regional agglomeration~\cite{novotny2014smart}. In such context, an interesting challenge is the taxi mobility prediction. Current advances in technology made available a variety of low-cost electronic devices, which can be of great support to rapidly move in the city traffic. Inside taxis, electronic GPS terminals are installed in order to grab the vehicle position or the taximeter status, and send this information to the dispatcher unit. In addition to GPS data, there are novel sources of information, e.g., data from mobile phones and Location Based Social Networks (LBSNs), that can be exploited to model the human mobility behaviour and to optimize the city traffic. Common LBSNs, like Foursquare\footnote{www.foursquare.com}, for instance, can provide the number and type of activities present in a target area (e.g., Arts \& Entertainment, Nightlife Spot, etc.), giving an insight on how many people can converge to that zone. All the collected information can then be used to infer the taxi destinations and the average travel time. 

In human mobility, the most straightforward way to predict the next location is to build a grid over an area of interest, then treating the problem as a multiclass classification, where the aim is to predict the next visited cell. For instance, in the winning approach \cite{de2015artificial} of the ECML/PKDD 2015 challenge\footnote{https://www.kaggle.com/c/pkdd-15-predict-taxi-service-trajectory-i}, the next taxi destination is obtained by employing a Multi-Layer Perceptron (MLP) network trained on the taxi trajectory, represented as a variable-length sequence of GPS points, and on diverse associated meta-information, such as the departure time, the driver identity, and the client information. Some of the limitations of previous approaches \cite{de2015artificial,yao2017serm} reside in the need of using the whole trajectory data and in the possibility of performing an instantaneous prediction of the destination only when most of the trajectory is available.

In this paper, we propose a Long-Short-Term-Memory network \cite{gers1999learning}, equipped with a self-attention module, to improve the prediction performance of the coordinates of the next drop-off point for a taxi. In particular, our model grounds on the individual driver's history, intended as the sequence of the last visited points, i.e., pick-up or drop-off points, instead of on the whole GPS trajectory. In order to represent such locations, we build a semantic representation based on LBSN data, coming from Foursquare. 
We tested our model on data referred to three different cities, namely Porto, as in the ECML/PKDD 2015 challenge, New York City (more specifically, Manhattan), and San Francisco. 
We have also compared our method with the winning model of the ECML/PKDD 2015 challenge under various settings, obtaining an improvement of 10.5\% (equal to approximately 0.355 km) in the average error distance. 

The rest of the paper is organized as follows. Section \ref{sec2} provides the state of the art on the taxi mobility and destination prediction tasks, whereas Section \ref{sec3} presents an in-depth description of the problem. In Section \ref{sec4}, we characterize the employed datasets and explain our methodological approach. Section \ref{sec5} delineates the proposed predictive model, while Section \ref{sec6} illustrates the experimental setup, together with the obtained results. Finally, conclusions and future research are drawn in Section \ref{sec7}.

\section{Related Work}
\label{sec2} 

Despite being far from adequately addressing the mobility management optimization problem, nowadays the variability of massive data describing human movements allows planners and policy-makers to face relevant urban challenges through computational methods \cite{gonzalez2008understanding,ahmed2016multi,barlacchi2017you,pappalardo2015returners}. While the observation of mobility flows offers the possibility of investigating the internal functioning of urban transport systems, this is not enough for an efficient transport planning. In this perspective, an interesting problem in taxi mobility is the optimization of the total profit guaranteed by the entire company fleet that comprises, for instance, the definition of a comprehensive mobility pattern able to help drivers to decide which of two potential rides will be more remunerative. Two interesting case studies are reported in \cite{phiboonbanakit2016does} and \cite{antoniades2016fare}, with respect to fare data related to Bangkok (collected by the Department of Land Transport of Thailand) and to New York City, respectively. A reinforcement learning approach to this task is also proposed in \cite{gao2018optimize}, whereas in \cite{santi2014quantifying} an analysis on the benefits of sharing a taxi ride, based on a shareability network, is carried out. In \cite{li2018analysis}, an interesting aspect of the path flow dynamics is illustrated, i.e. that the shortest path between the origin and the destination is not always the best option for a taxi driver who, in 90\% of the cases, choose instead on personal, knowledge-based optimality criteria. In a follow-up study of \cite{santi2014quantifying}, a network-based solution to the minimum fleet problem is provided, obtained by investigating the minimum number of vehicles needed to serve all the trips without incurring in significant delays \cite{vazifeh2018nature}. The method is tested on a data set of 150 million taxi trips taken in the city of New York over one year. The real-time implementation of this approach, gaining near-optimal service levels, allows a 30\% reduction compared to common fleet sizes. Such improvements follow from a mere re-organization of the taxi dispatching, without assuming ride sharing or requiring changes to taxi regulations.

Some other research works investigated the taxi mobility also in terms of selecting the optimal pick-up point and the route to find new customers \cite{hu2012modeling}. For instance in \cite{yuan2011find}, different strategies to find new taxi customers are presented. In this case, taxi trajectories are used to predict if it is preferable to wait in an ad-hoc parking area or to travel without passengers. Also in \cite{smith2017predicting}, a Random Forest regression approach is adopted to predict the taxi pick-up zones in New York City (NYC), based on start/end points of trajectories and Twitter data. In \cite{wong2014modelling} is proposed an in-depth analysis of vacant trip routes, while \cite{wong2014bi} implement a two stage decision process for predicting if a driver will reach the nearest taxi stands and to state whether the drivers will join the taxi queue at the taxi stands. Also in \cite{tang2016two} is shown a two stage prediction model, inferring the driver's pickup location together with the choosed route. A recommendation model for both the driver and the customer is devised in \cite{yuan2011find}, reducing the distance covered without any passenger, and the waiting time, respectively. In \cite{li2012prediction}, an auto-regressive integrated moving average (ARIMA) method is built to forecast the spatio-temporal variation of passengers in a hotspot, discovering patterns of pick-up quantity (PUQ) for urban hotspots. Moreover, a rich literature exists on different implementations of Hidden Markov Models for mobility analysis and optimization \cite{simmons2006learning,krumm2016markov}. For instance, in \cite{chen2014nlpmm,gambs2012next} the probability of the future displacement is computed through a transition matrix between city hotspots, obtained by the past trajectories, whereas in \cite{monreale2009wherenext} the next location is estimated, based on GPS data, by applying trajectory pattern mining. Finally, a variety of neural network approaches have been employed to model the problem of next location prediction. For instance, \cite{liu2016predicting,du2016recurrent,yao2017serm,feng2018deepmove} present interesting solutions to capture the regularities characterizing human mobility based on taxi rides, while in \cite{lv2018t} trajectories are modeled as two-dimensional images that constitute the input for a Multi-Layer Convolutional neural network used to predict the next destination point.

\section{Background}
\label{sec3}
In this section, we provide a formal definition of the taxi destination prediction problem, introducing the necessary notation and background theory. Then, we briefly introduce the Recurrent Neural Network and the attention mechanism. Finally, we discuss motivation and implications of our proposed solution.

\subsection{Notation and problem definition}

\begin{definition}
A \textit{spatio-temporal point} $p$ is defined as the couple $p=(t, l)$, where $t$ indicates the time at which the location $l=(x, y)$ is visited, being $x$ and $y$ spatial coordinates in a given coordinate reference system (CRS), e.g. latitude and longitude in the CRS WGS84. 
\end{definition}

In our framework, we aim at modeling the taxi drivers' behaviour. To this extent, it is useful to introduce the definition of what we call a \textit{taxi trajectory}.

\begin{definition}
\label{trj_simple}
Let $u$ be a taxi driver. A \textit{taxi trajectory} $T_u=p_{1}, p_{2}, ..., p_{k}$ is a time-ordered sequence composed by alternating pick-up and drop-off spatio-temporal points, describing the last $k/2$ taxi rides of driver $u$.
\end{definition}

As in the ECML/PKDD 2015 challenge, we focus on predicting where a taxi will drop-off a client.

\begin{problem}
\label{pr:destination}
Let $u$ be a taxi driver and $T_u$ his/her taxi trajectory.
Given $T_u \cup p_{k+1}$, being $p_{k+1}$ the current pick-up point, we define the \textit{next taxi destination task} as the problem of predicting the drop-off location $l_{k+2} = (x_{k+2}, y_{k+2})$, which will be the actual destination of the driver $u$.
\end{problem}

In the following, we describe a Recurrent Neural Network (RNN) architecture aimed at modeling the behaviour of each taxi driver during a sequence of pick-up and drop-off points, in order to predict the coordinates of the next drop-off point.

\subsection{Recurrent Neural Networks and Long-Short-Term-Memories}
Recurrent Neural Networks are equipped with feedback connections that produce internal loops. Such loops induce a recursive dynamics within the networks and thus introduce delayed activation dependencies across the processing elements. In doing so, RNNs develop a kind of memory, that makes them particularly tailored to process temporal (or even sequential) data, e.g., coming from written text, speech, or genome and protein sequences. Unfortunately, properly training RNNs is hard, due to both the vanishing and the exploding gradient pathologies, which introduce the long-term dependency problem, making difficult the processing of long sequences.

Long-Short-Term-Memory networks --- introduced in \cite{hochreiter1997long} and usually just called LSTMs --- are a special kind of RNNs, capable of learning long-term dependencies. All these networks have the form of a chain of repeated modules where each module is composed by four interacting layers with different functions.

In particular, the forget layer $f_t$ selects the part of the cell state $h_{t-1}$ which is responsible for removing the information no longer required for the LSTM to carry on its task. This allows the optimization of the LSTM performances. The forget gate processes also $x_t$, i.e. the input at the current time step. Instead, the input gate $i_t$, whose elaboration depends on $h_{t-1}$ and $x_t$, is responsible for the addition of information to the cell state at time $t$; next, a \textit{tanh} layer creates a vector of new candidate values, $\tilde{C}_t$, possibly added to the cell state. Then, these two sources of information are combined to create an updated state. Finally, the output gate $o_t$ selects those parts of the cell state that must be produced in output. The complete computing algorithm for LSTMs can be summarized as follows:
\begin{eqnarray}
    \begin{aligned}
        &f_t=\sigma(W_f[h_{t-1},x_t]+b_f)\\
        &i_t=\sigma(W_i[h_{t-1},x_t]+b_i)\\
        &\tilde{C}_t=tanh(W_C[h_{t-1},x_t]+b_C) \\
        &C_t=f_t*C_{t-1}+i_t*\tilde{C}_t\\
        &o_t=\sigma(W_o[h_{t-1},x_t]+b_o) \\
        &h_t=o_ttanh(C_t)
    \end{aligned}
\end{eqnarray}

\subsection{The attention mechanism}
One of the most interesting human cognitive processes is the attention mechanism, a well studied phenomenon also in neuroscience. Indeed, just looking at an image, humans are able to focus on certain regions with \emph{high resolution}, while perceiving the surrounding parts as \emph{background}. In this way it is then possible to adjust the focal point over time. A very similar approach can be attached to deep learning tasks --- even if attention in neural networks is, indeed, loosely related to the visual attention mechanism found in humans. 

The main idea behind \emph{attention} is to produce a score for each element of the current input (in the sequence case a score for each timestamp of the sequence itself). The implementation of this mechanism is quite straightforward and suited to this case study. Formally speaking, an attention model is a method that takes $n$ arguments $y_1, \ldots, y_n$ and a context $C$, and returns a vector $a$ which is supposed to be the summary of the inputs, focused on particular information dictated by the context $C$. This can be obtained as a weighted arithmetic mean of $y_i,\, i=1,\ldots,n$, with the weights chosen according to the relevance of each $y_i$, given the context $C$.

Attention mechanisms have been applied to a very wide range of applications, such as speech recognition \cite{chorowski2015attention}, machine translation \cite{bahdanau2014neural}, text summarization \cite{rush2015neural}, and image description \cite{xu2015show}.
They are widely used also in few-shot learning \cite{mishra2018simple,vinyals2016matching}, in order to compare an unknown sample and a set of labeled samples called the support set, indeed focusing the attention on the most similar support samples to the queried one. Finally, attention has been successfully employed for human mobility \cite{feng2018deepmove}, in order to capture the important periodicities that govern human movements \cite{gonzalez2008understanding,schneider2013unravelling}. However, this approach normally rely on historical GPS traces, which are often not available in the case of taxi data and to overcome this problem we propose to apply the attention mechanism to the taxi trajectory itself (see Definition \ref{trj_simple}).  

\section{Data description}
\label{sec4}
Taxi trajectory data are of great interest for the observation, evaluation, and optimization of transportation infrastructures and policies. For example, major problems of modern cities, such as traffic jams, are caused by an improper road planning, maintenance, and control. Taxi trajectory datasets are available nowadays for many  large cities \cite{zheng2011urban,santi2014quantifying} and they provide a valuable resource for modeling and understanding urban transportation patterns and human mobility behaviours. In our study, we focus on Porto, Manhattan (New York City)\footnote{https://chriswhong.com/open-data/foil\textunderscore nyc\textunderscore taxi/}, and San Francisco \cite{epfl-mobility-20090224}. In particular, the Porto dataset was released in the context of the ECML/PKDD 2015 challenge and hosted as a Kaggle competition\footnote{https://www.kaggle.com/c/pkdd-15-predict-taxi-service-trajectory-i}, which allows us to easily compare our model with the winner's approach. 
In order to provide semantic information about people's activities in each visited location, we enriched the taxi dataset with geo-located texts. In particular, we provide a feature representation of each location in terms of Foursquare Point-Of-Interests (POIs). We used freely-available data sources in order to make our research and experiments easily reproducible.

\subsection{Taxi Mobility Data}
\label{taxi_data}
\begin{figure}[H]
    \centering
    \begin{subfigure}
            \centering
            \includegraphics[width=0.45\textwidth]{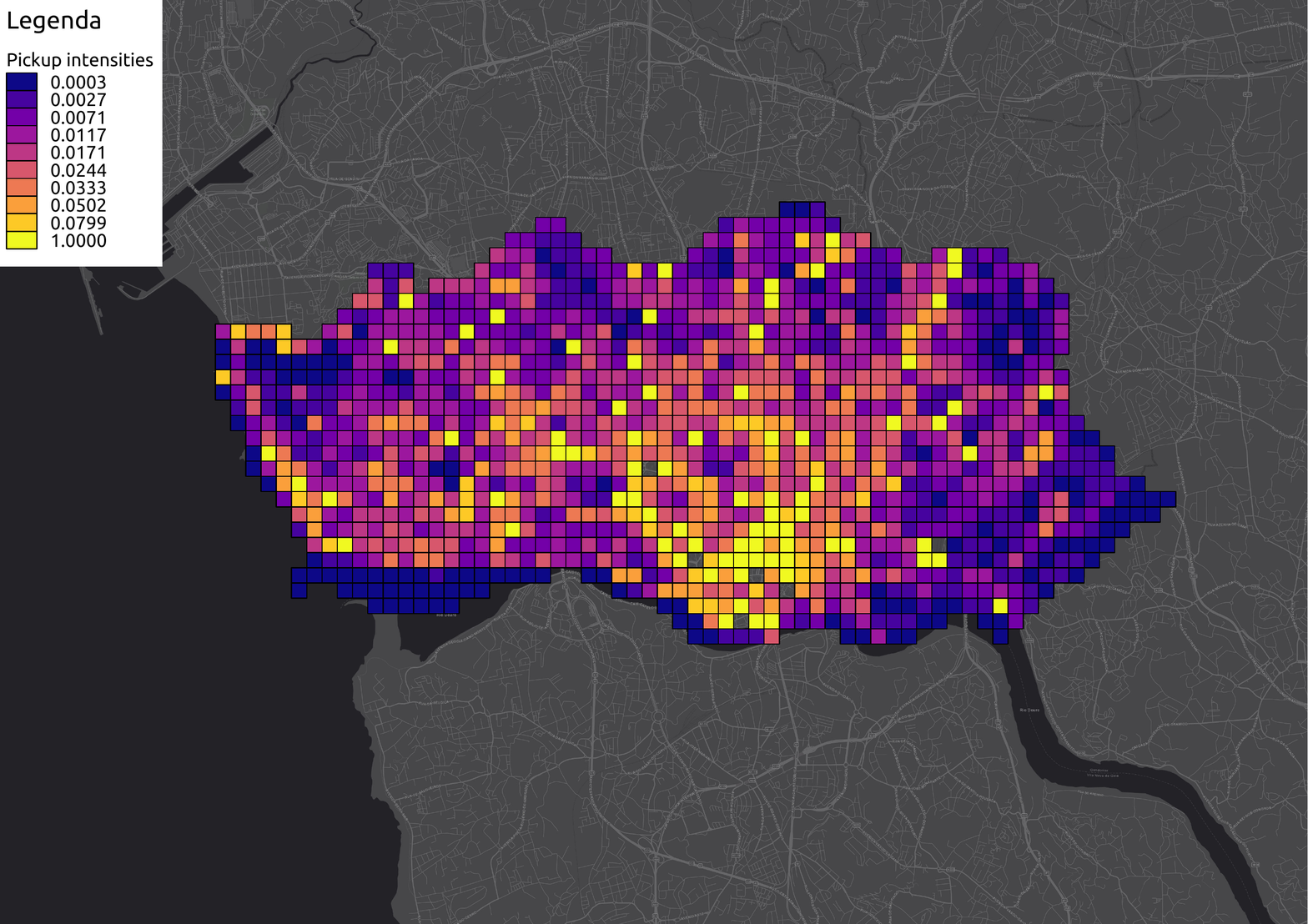}
    \label{fig:pickup}
    \end{subfigure}
\begin{subfigure}
            \centering
            \includegraphics[width=0.45\textwidth]{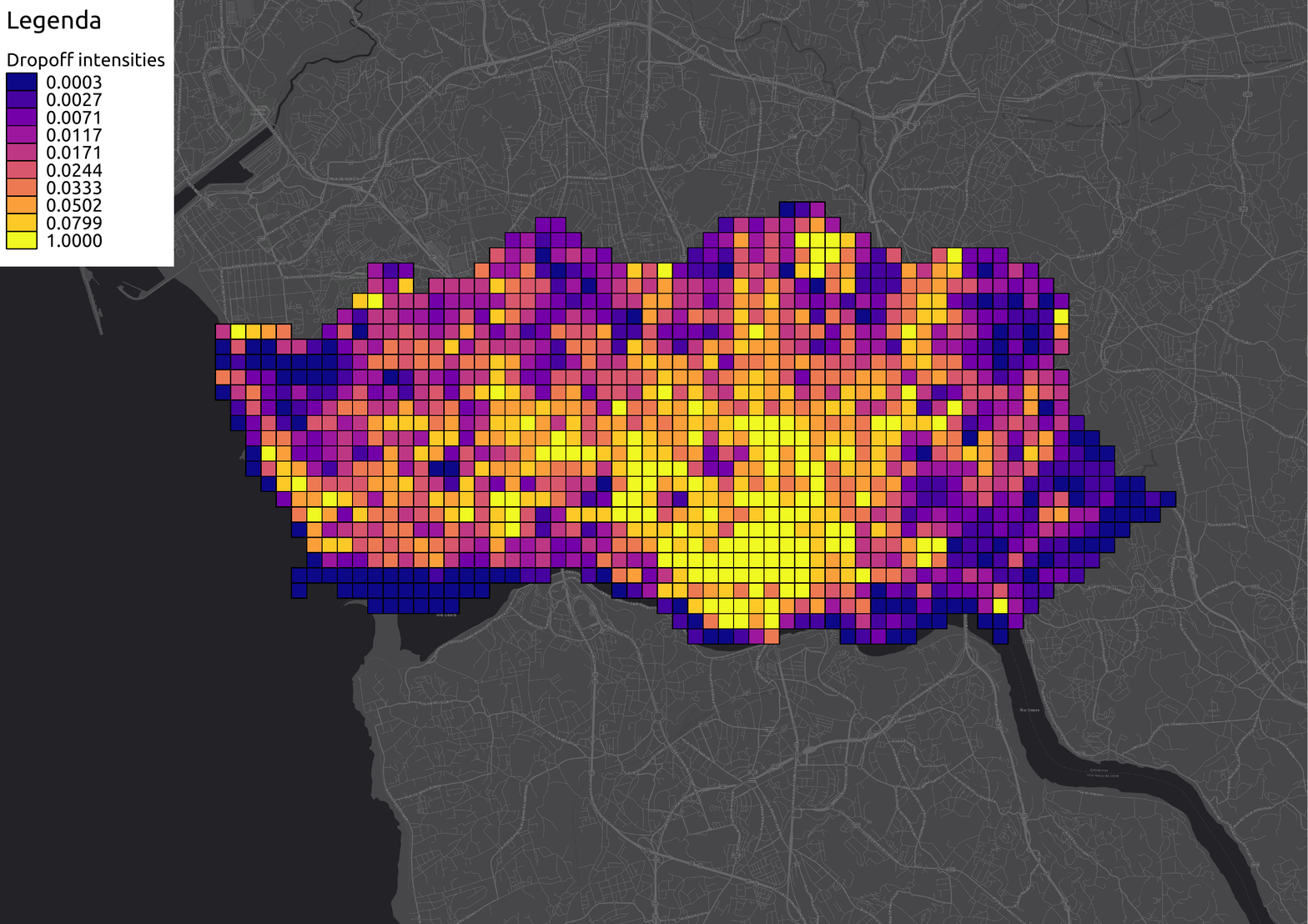}
    \label{fig:dropoff}
    \end{subfigure}
    \caption{Pick-up (left) and drop-off (right) intensities in the city of Porto. Data are linearly scaled between 0 and 1.}
\label{fig:distr}
\end{figure}

The Porto taxi dataset is composed by 1.7 million records, coming from 442 taxis running in Porto, for a period ranging from 2013-07-01 to 2014-06-30. Figure \ref{fig:distr} reports the pick-up and drop-off distributions in Porto. For each ride, a GPS trace is provided, making the description of a taxi trajectory different from the one introduced in Definition \ref{trj_simple}. In this case, the first point represents the pick-up location, the last one is the drop-off place, while in between there are spatio-temporal points sampled every 15 seconds. In addition, other metadata, like the \textit{taxi id}, the \textit{type} and the \textit{origin of the call}, the \textit{day type} (i.e., holiday, working day, weekend), and the \textit{starting ride timestamp}, are attached to each trip. Such information is useful in order to detect some recurrent patterns, such as the flow of people who go to work or return home in specific time slots on working days. Similarly, for San Francisco, we can rely on trips of 536 taxis. The dataset contains GPS trajectories for 464019 trips in 2008. The trajectories, as in the Porto dataset, describe the whole trip with points sampled every 10 seconds. 
The last, and biggest, dataset contains data for Manhattan and provide taxi trajectories defined as an Origin-Destination matrix. It is composed by 13426 taxis, driven by 32224 different drivers, during the entire 2013. Each month accounts for approximately 15 million trips. We selected the first three months of 2013, considering the most 5994 active drivers, which turns out in a total of 9362829 trips. All the above mentioned datasets do not contain vacant trips, which means that all the trips are made with a customer on board. In addition to the trajectory data, both these datasets also provide the driver's id and the starting time of the ride. 

Following Definition \ref{trj_simple}, for both San Francisco and Porto, we map each GPS trace of the same taxi driver into a sequence of pick-up and drop-off points. More in detail, we group the trips by the driver's id, and then we sort them in ascending order based on their timestamp. Then, for each trace we select the pick-up and the drop-off point and create a taxi trajectory. In particular, given a set of taxi trajectories as in Definition \ref{trj_simple} and the problem setting as in Problem \ref{pr:destination},\ we impose $k=8$ in order to keep up to four trips from the past, i.e., we use up to four pairs of pick-up and drop-off points. 
Such a choice represents a good trade-off between keeping a relevant past history while maintaining a sufficient number of trajectories for each driver and thus enabling the model to learn the cabdriver habits. Finally, in order to select trips relative to the same driver work shift, we use trips that are not separated by more than three hours each other. 

Following this procedure, we obtain 260600 trajectories for Porto, 87548 for San Francisco, and more than a million for Manhattan. In the last case, we select 600 drivers out of the initial 5994, for computational reasons, obtaining 184000 sequences.

\subsection{Points Of Interest}\label{pois}
A Point Of Interest (POI) is usually characterized by a location (i.e., latitude and longitude), some textual information (e.g., a description of the activity in that place), and a hierarchical categorization, that provides different levels of detail about the activity of a particular place (e.g., \emph{Food}, \emph{Asian Restaurant}, \emph{Chinese Restaurant}). 

We chose to use the Foursquare API\footnote{https://developer.foursquare.com/} as a source of POIs. Foursquare is a geolocation-based social network supported with web search facilities for places and recommendation systems, accounting for 100,000 free of charge API requests per day. Every LBSN has its own hierarchy of categories, which is used to characterize each location and activity (e.g., restaurant or shops) in the database. Thus, each POI is associated with a hierarchical path, which semantically describes the type of location/activity. For example, for the activity \emph{Chinese Restaurant}, the path \textit{Food} $\rightarrow$ \emph{Asian Restaurant} $\rightarrow$ \emph{Chinese Restaurant} is established. This path is much more informative than just the target POI name, as it provides feature combinations following the structure and the node proximity information. For each POI we consider only the macro-category and, in particular, Foursquare provides ten different macro-categories, namely \textit{Arts and Entertainment, College and University, Event, Food, Nightlife Spot, Outdoors and Recreation, Professional and Other Places, Residence, Shop and Service, Travel and Transport.}

We extracted 8928 POIs for Porto, 72567 for Manhattan, and 30059 for San Francisco. Despite the availability of many other interesting geographical datasets, e.g., land use and census data, we decided to limit our model by only using POIs data. This is due to the fact that those additional geographical datasets are not consistently defined among different cities, making their availability not uniform in terms of definition and coverage.

\section{Learning to Predict the Next Taxi Destination}
\label{sec5}
A mobility trace is a temporally-ordered collection of GPS locations. As the majority of sequential data, mobility traces are suited to be treated with models such as Recurrent Neural Networks (RNNs) instead of static architectures like Multi Layer Perceptrons (MLPs), which are not tailored to work with temporal and sequential data.  

Previous approaches for predicting the next taxi destination were mainly focused on fine-grained single trajectories, meaning that they were based on the whole GPS trace of each ride \cite{de2015artificial}, showing some important limitations such as a huge amount of data to be stored (i.e., all the GPS points of a trip), and the need of knowing almost all the trajectory in order to predict its final point.
Instead, to avoid the aforementioned problems, we modeled the trajectory as in Definition \ref{trj_simple}. It turns out a sequence composed by pairs of GPS points (pick-up and drop-off locations) of several taxi rides. In this way, there is no need to wait the whole trajectory and the prediction can be performed instantaneously as soon as the trip starts and the pick-up location is available. Finally, in most of the mobility models, the next location prediction is set as a classification problem, where the goal is to classify to which locations, belonging to a known set, our driver is moving \cite{yao2017serm,feng2018deepmove}. The main drawback of this approach is that many locations will never be produced by the model, namely unseen locations in the training set. To overcome this limitation, we propose to predict the coordinates of the destination, by approximating directly two different functions, for the latitude and the longitude, respectively.

\begin{figure}[htb!]
\label{cube}
\centering
\includegraphics[width=5cm, height=5cm]{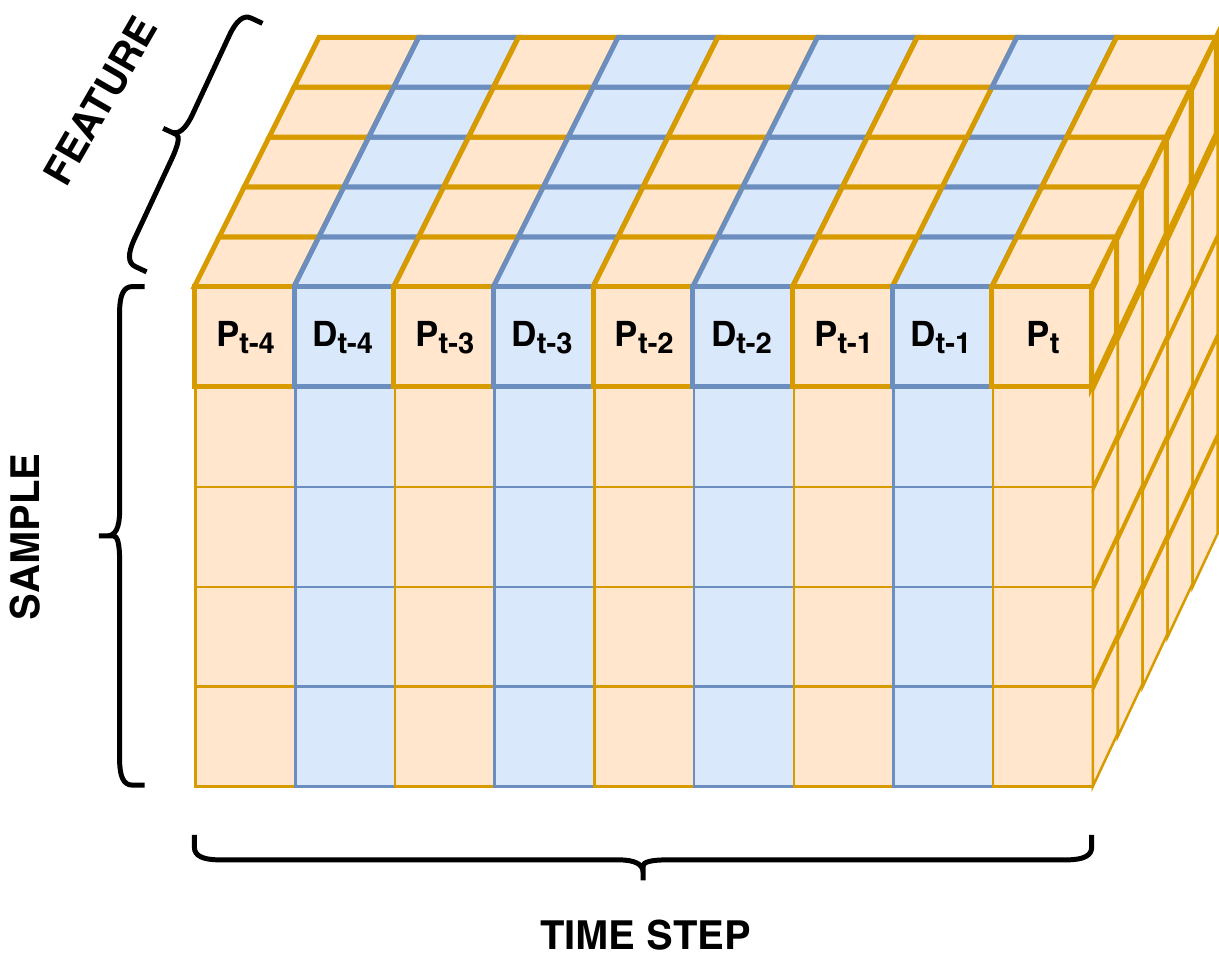}
\caption{The driver's trajectory fed into our model (\textit{P} marks the \textit{pick-up} points and \textit{D} marks the \textit{drop-off} points).}
\label{t_set}
\end{figure}

The adopted RNN architecture is relatively simple: it uses a Recurrent Neural Network in order to model the taxi behaviour, looking at the trip history and at the geographical information, to include the semantic meaning of each visited location. The driver's trajectory fed into our model is sketched in Fig. \ref{t_set}. 

\subsection{General framework}
\begin{figure}[htb!]
\label{net}
\centering
\includegraphics[width=8cm, height=8cm]{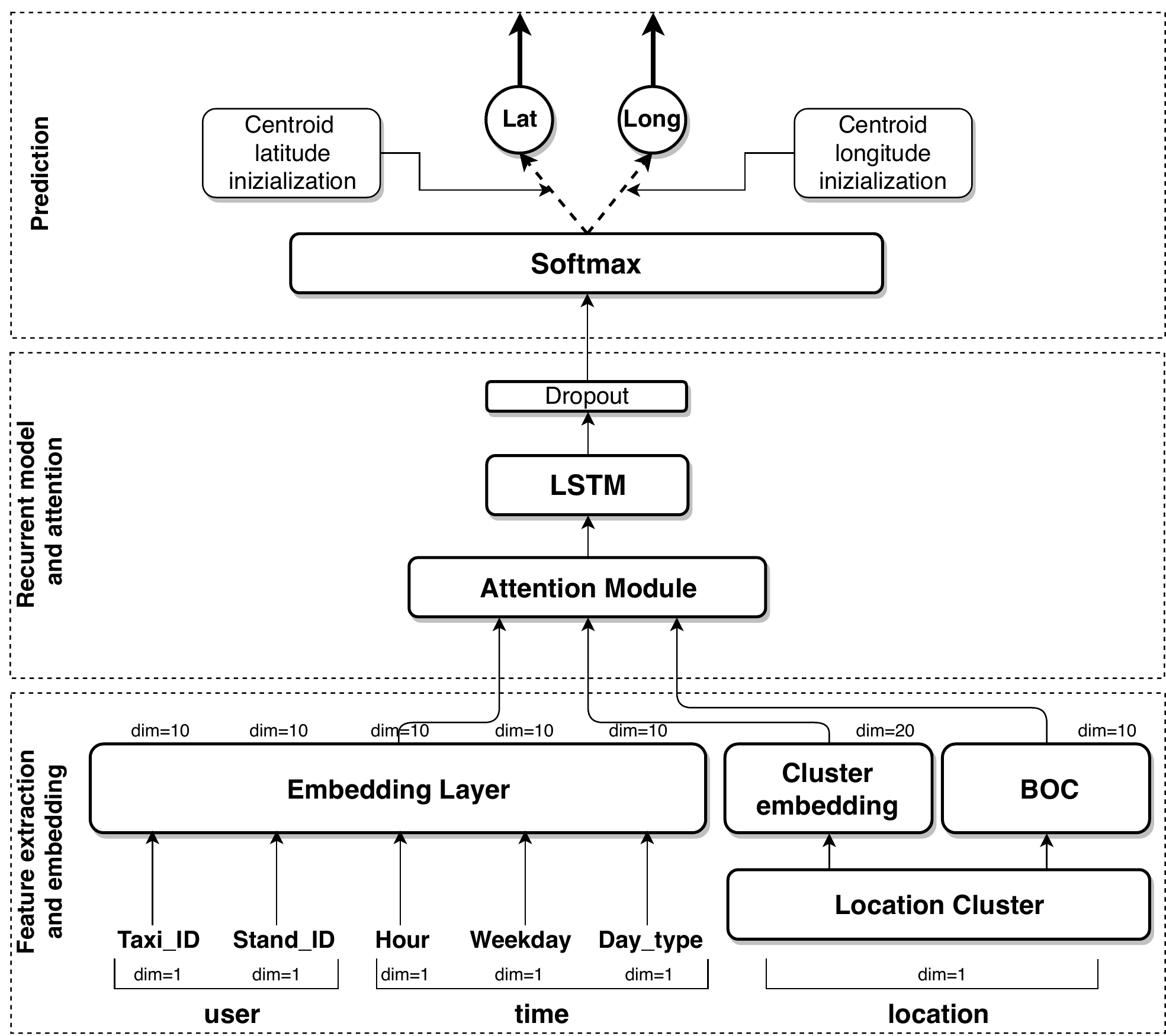}
\caption{The architecture of our model with the three submodules: (i) Feature Extraction and Embedding, (ii) Recurrent Model and Attention, and (iii) Prediction.}
\label{fig:arch}
\end{figure}

Our framework consists of three main components: (i) Feature Extraction and Embedding, (ii) Recurrent Module with Attention, and (iii) Prediction. The proposed neural network architecture is illustrated in Fig. \ref{fig:arch}.

First, we define a set of location clusters $C$ with $m=\vert C \vert$, calculated with a clustering algorithm, i.e., K-means, on the destinations of all the training trajectories. Each point of the trajectory is consequently assigned to the closest centroid $c_i$. In this way, the trajectory $T_u$ is translated into a new mapped cluster trace $CT_u=c_1, c_2, ..., c_k$, where $c_i$ is the hash of the closest cluster to the point.
Since we are working with latitude and longitude values, we use the \textit{haversine distance} to compute the distance between two points, $p_1$ and $p_2$:

\begin{eqnarray}
d_{haversine}(p_1,p_2)=2R\left(\sqrt{\frac{a(p_1,p_2)}{a(p_1,p_2)-1}}\right) \\
\begin{split}
a(p_1,p_2)=sin^2\left(\frac{\phi_2-\phi_1}{2}\right)+cos(\phi_1)cos(\phi_2)sin^2\left(\frac{\lambda_2-\lambda_1}{2}\right)
    \end{split}
\end{eqnarray}

\noindent
where $R$ is the earth radius and $(\lambda_i,\, \phi_i), \, i=1,2$ are the longitude and the latitude of $p_i$, respectively. 

We model each location with an embedding layer in order to capture all the factors that influence the human mobility, such as time, day and the semantic characterization of each place. In addition, we represent each location as a single word and we apply Word2Vec \cite{mikolov2013efficient} on the resulting sequences, which allows us to obtain a dense representation based on the co-occurrence of both the origin and the destination. Then, based on a recurrent module, we collect all the sequential information we can derive from the past visited locations. For this task, we employ an LSTM architecture as the basic recurrent unit, because of its effectiveness in modeling human mobility \cite{liu2016predicting}. Following the idea proposed in \cite{feng2018deepmove}, we also apply an attention mechanism on the input sequence, in order to capture mobility regularities from previous visited locations. Finally, a prediction module composed by a softmax layer, followed by a linear layer, is designed to estimate the latitude and the longitude values of the next taxi destination.

\subsection{Feature extraction and embedding}\label{embedding}

Mobility transitions are governed by multiple factors, such as the day-time and the geographical characteristics of visited places. In particular, we can assign to each location a \textit{particular meaning} related to the activities present in that zone. Thus, an effective representation of a particular area turns out to be very important for mobility data analysis since it allows to enrich the information related to each location. We propose a multimodal embedding module to jointly embed the spatio-temporal and the driver features into a vector representation, which is then used as the input for the RNN-based prediction model.

\subsubsection{Driver Behavioral Features}
We represent the behaviour of the driver with a set of categorical features: (i) the time, expressed with the hour of the day $h \in [0,23]$; (ii) the week-day $wd \in [0,6]$; and (iii) the day-type, $dt \in [0,2]$ (i.e., workday, pre-holiday, and holiday). Instead of modeling these categorical features with the simple one-hot representation, we follow the approach of producing a dense representation vector \cite{mikolov2013efficient} based on an embedding layer, whose weights are updated during the training phase.

\subsubsection{Spatial Semantic Features}
Given the set of location clusters $C$ and a set of POIs, we assign each POI to the closest cluster. Thus, for each point of a given mapped trajectory $CT_u$, we build a feature representation making use of the associated POIs. Every venue is hierarchically categorized (e.g., \textit{Professional and Other Places $\rightarrow$ Medical Center $\rightarrow$ Doctor's office}) and the categories are used to produce an aggregated representation of the area. We model spatial semantic features by using the Bag-Of-Concepts (BOCs) representation proposed in \cite{barlacchi2017structural}, based on aggregating all the POIs associated to the trajectory and counting their macro-categories (described in Section \ref{pois}, e.g., \textit{Food}). In other words, Bag-Of-Concept (BOC) features are generated by counting the number of activities for each category in a cluster. 

\subsubsection{Spatial Zone Embedding}
Spatial Semantic Features provide a time-independent representation, which does not capture the \textit{dynamics} of a particular area. In this perspective, we can use human mobility data to learn a dense representation based on the mobility flows in the city. In the resulting space, embeddings of zones with similar urban mobility will result geometrically close each other. Human mobility can be seen as a language, where sequences of locations are sequences of words. It turns out that sequence models designed for textual data, and used in natural language processing (NLP), can be properly applied to a sequence of locations. For instance, Word2Vec \cite{mikolov2013efficient} is a well-known textual embedding technique to learn word embeddings. Given a word, the embedding is learnt from the co-occurrence words belonging to a near by word-window. In this way, in the generated space of Word2Vec, two semantically similar words will results to have a similar vector representation. Given the set of location clusters $C$, we assign a textual label to each cluster. Then, we map each trajectory in a sequence of words, enabling Word2Vec to learn the zone embedding by relying on the mobility relation from different zones.

\subsection{Recurrent module with attention}
\begin{figure}[htb!]
\centering
\includegraphics[width=3.5cm, height=5cm]{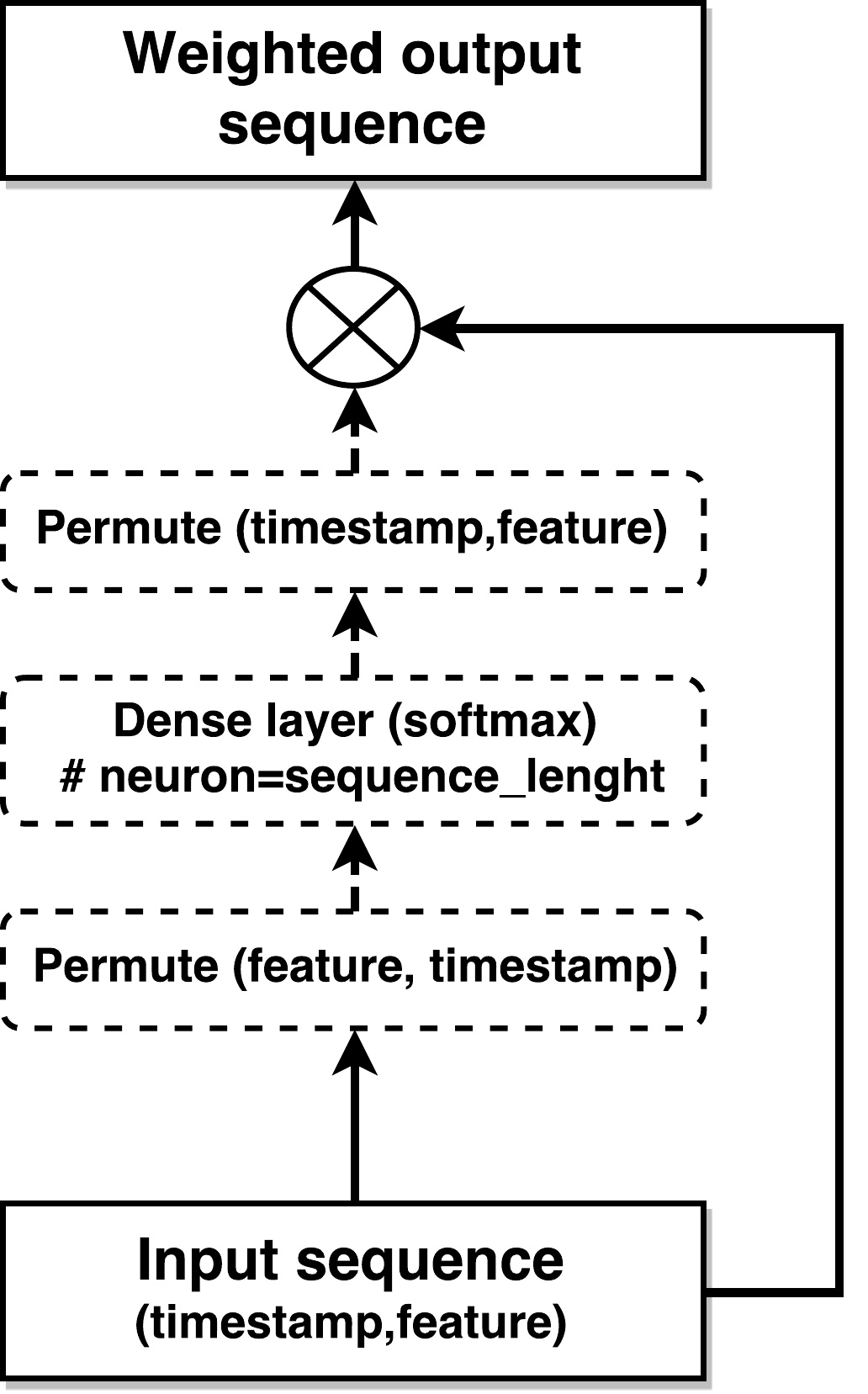}
\caption{The attention mechanism applied on the input trajectory.}
\label{fig:attent}
\end{figure}

Recurrent Neural Networks are particularly suited to process sequential data, such as GPS traces \cite{liu2016predicting,yao2017serm,feng2018deepmove}. For this reason, we model our problem using Recurrent Neural Networks with an LSTM as a basic unit, having ReLu as the activation function. First, we apply the attention mechanism to the spatio-temporal vector sequence in input. The implemented structure is presented in Fig. \ref{fig:attent}. 
The aim of the attention module is to learn on which part of the trajectory is more important to focus on. This is done by transposing the input and feeding it to a softmax, which estimates the weight distributions, that are then combined with the input sequence. Thus, the recurrent layer is focused on specific parts of the sequence, to capture mobility regularities from the current trajectory. Finally, a dropout layer is applied before the softmax layer, in order to prevent overfitting.

\subsection{Prediction}
The prediction module represents the last component of our architecture, that combines the output of the previous modules and completes the prediction task. In particular, it consists of a softmax layer and a linear layer. First, the softmax layer, having $m=|C|$ as the number of neurons, takes the representation generated by the recurrent module as its input. Then, since the network must evaluate the coordinates of the destination point, an additional output layer with two neurons is added, representing the latitude and the longitude coordinates, respectively. It is worth noting that this operation is equivalent to add a simple linear output layer, whose adjustable weight matrix is initialized with the cluster centers. 
Thus, the output of this layer is defined as:
\begin{equation}
    \begin{aligned}
    &\Tilde{y}=\sum_{i=1}^{C}P_ic_i \\
    &P_i = \frac{exp(e_i)}{\sum_{j=i}^{C}exp(e_j)}
    \end{aligned}
\end{equation}
\noindent
where $P_i$ is the softmax probability associated to each cluster point.

\section{Experimental results}
\label{sec6}
The following experiments aim at demonstrating the effectiveness of our model for predicting the next destination in a taxi ride. We experimented our approach on three different taxi datasets for the cities of (i) Porto, (ii) Manhattan (New York City), and (iii) San Francisco, also enabling comparisons with state-of-the-art models \cite{du2016recurrent}. 
\subsection{Experimental setup} \label{exp_setup}

We performed experiments in Porto, Manhattan and San Francisco by using 260.600, 184.000 and 87.500 taxi rides, respectively. For the three cities we used the whole datasets, randomly splitting the data into 65\%-15\%-20\% for the training, validation, and test sets. For the city of Porto and San Francisco the available dataset is composed by the complete trajectories, while for Manhattan the trajectories are only composed by pick-up and drop-off points, as in Definiton \ref{trj_simple}). Due to the unbalancing of the datasets and the spatial nature of the problem, standard classification measures such as accuracy and F1-score are not appropriate, not giving an adequate quantification of the error.
Hence, we used the Error Distance Score (EDS), which is defined as the Haversine distance between the predicted point and the actual destination of the trip: 
\begin{equation}
    EDS = d_{haversine}(\Tilde{y}, y)
\end{equation}
where $\Tilde{y}$ is the predicted point and $y$ is the correct destination. 
 
The experimental study compares our work with several baseline approaches and state-of-the-art models, listed in the following:
\begin{itemize}
\item \textbf{Nearest Neighbors (NN)}: Given a pick-up point, it outputs the coordinates of the closest cluster centroid. 
\item \textbf{MMLP}: The Multi-Layer Perceptron (MLP) implementation that won the ECML/PKDD 2015 challenge \cite{de2015artificial}. In this model, the input layer receives a representation of the taxi trajectory, composed by the first five and the last five GPS points, and its associated metadata. The network is composed by a standard hidden layer, containing 500 Rectifier Linear Units (ReLUs) \cite{glorot2011deep}, followed by a softmax layer with $m=|C|$ neurons. As in our approach, each coordinate is assigned to a given set of cluster centroids. Thus, the output layer predicts $\Tilde{y}$, which is a weighted average between softmax output and the destination cluster centroids. The optimization problem is modeled as a multi-class classification, having the cross-entropy as the loss function. The network is trained with Stochastic Gradient Descent with a batch size of 200, a learning rate of 0.001 and a momentum term of 0.9. To perform a fair comparison, we trained the MMLP model on the last trip of our trajectory (Definition \ref{trj_simple}), thus obtaining the same number of training, validation and test patterns.  
\item \textbf{MMLP-SEQ}: It is trained following the same approach of MMLP but, instead of using the first five and the last five GPS points as input, we have linearized the driver's trajectory (Definition \ref{trj_simple}). 
\end{itemize}

\subsection{Training and hyperparameter setting}
We tuned the parameters of our models for each city by performing a grid search on the number of neurons, on the number of layers and on the learning rate. The performance was evaluated on the validation set and the selected values are reported in Table \ref{table1}.
The K parameter used in the K-means algorithm is set to 3392 for Porto, as in the original work \cite{de2015artificial}, to perform a fair comparison, and to 2000 for Manhattan and San Francisco, being both areas smaller than the extension of Porto. Further investigations on the number of clusters could be of interest but are out of the scope of this paper. For the embedding of both the driver and the time features, we adopted a layer of size 10, for each single feature, as in \cite{de2015artificial}. The size of the location embedding is 20, while the BOC representation has 10 components, equal to the number of the macro-categories present in Foursquare. The input dimension of the LSTM (with ReLu activation functions \cite{glorot2011deep}) is equal to the size of the vector obtained after concatenating all the previous representations. To train the network, we used the Adam optimizer \cite{kingma2014adam}, approximating the latitude and the longitude values separately, but sharing the underlying structure --- embeddings, attention module and LSTM layer. The network is trained using the Mean Squared Error (MSE) as the loss function and implementing the early stopping. This means that we stop the training procedure if no changes in the MSE occur on the validation set for at least 10 epochs. We compute the MSE score on the validation set after each epoch and save the network parameters if a new best MSE score is obtained. During the test phase, we use the parameters of the network that produced the best MSE score on the validation set.  The dropout rate is set to $p = 0.5$. The word embedding for the textual label of each cluster is done by using the Gensim\footnote{https://radimrehurek.com/gensim/} implementation \cite{rehurek2010software}. We opted for a Continuous Bag-of-Words (CBOW) model with a window size of 5 and without filtering words on frequency. The dimensionality of this embedding is set to 20.

\begin{table}[H]
\centering

\begin{tabular}{|l|c|c|c|}
\hline
\textbf{City}  & \textbf{LSTM Neurons} & \textbf{Learning Rate}  & \textbf{Batch Size} \\ \hline
\textbf{Porto} & 128 & $10^{-3}$ & 64 \\ \hline
\textbf{San Francisco} & 128 & $10^{-3}$ & 64  \\ \hline
\textbf{Manhattan} & 256 & $10^{-3}$ & 64       \\ \hline
\end{tabular}
\vspace{1.0em}
\caption{Training parameters for each city.}
\label{table1}
\end{table}

\subsection{Results}
The performances of the compared approaches are listed (per column) in Table \ref{result}. We indicate with LSTM the simplest approach that uses the input representation only constituted by the driver and the time representation. The zone, i.e., the coordinates of the cluster location, is fed into an embedding layer with randomly initialized weights, allowing their update during the training. LSTM (BOC), indicates the model with the BOC features in input (see Section \ref{embedding}). Finally, LSTM (BOC+W2V) represents the Recurrent Neural Network that includes both the BOC features, but with the representation of the zone embedding obtained with (W2V).

\begin{table}[H]
\centering
\caption{Error Distance Score (km) for Porto, San Francisco and Manhattan. $^{(*)}$ The trajectory is composed by pick-up and drop-off points only.}
\label{result}
\begin{tabular}{|l|c|c|c|}
\hline
\multicolumn{1}{|c|}{\textbf{Model}}                                 & \multicolumn{1}{l|}{\textbf{Porto}} & \multicolumn{1}{l|}{\textbf{San Francisco}} & \multicolumn{1}{l|}{\textbf{Manhattan}} \\ \hline
\textbf{NN}                                                          & 3.215                               & 3.023                                       & 2.375                                   \\ \hline
\textbf{MMLP}                                                        & 3.211                               & 1.994                                       & 2.543$^{(*)}$                                  \\ \hline
\textbf{MMLP-SEQ}                                                    & 3.003                               & 2.762                                       & 2.554                                   \\ \hline
\textbf{LSTM}                                                        & 2.923                               & 2.547                                       & 2.111                                   \\ \hline
\textbf{\begin{tabular}[c]{@{}c@{}}LSTM (BOC)\end{tabular}}        & 2.923                               & 2.397                                       & \textbf{2.085}                                   \\ \hline
\textbf{\begin{tabular}[c]{@{}c@{}}LSTM (BOC+W2V)\end{tabular}} & \textbf{2.88}                      & \textbf{2.270}                              & 2.088 \\ \hline
\end{tabular}
\end{table}
\begin{figure}[H]
\centering
\subfigure{
	\includegraphics[width=.31\textwidth]{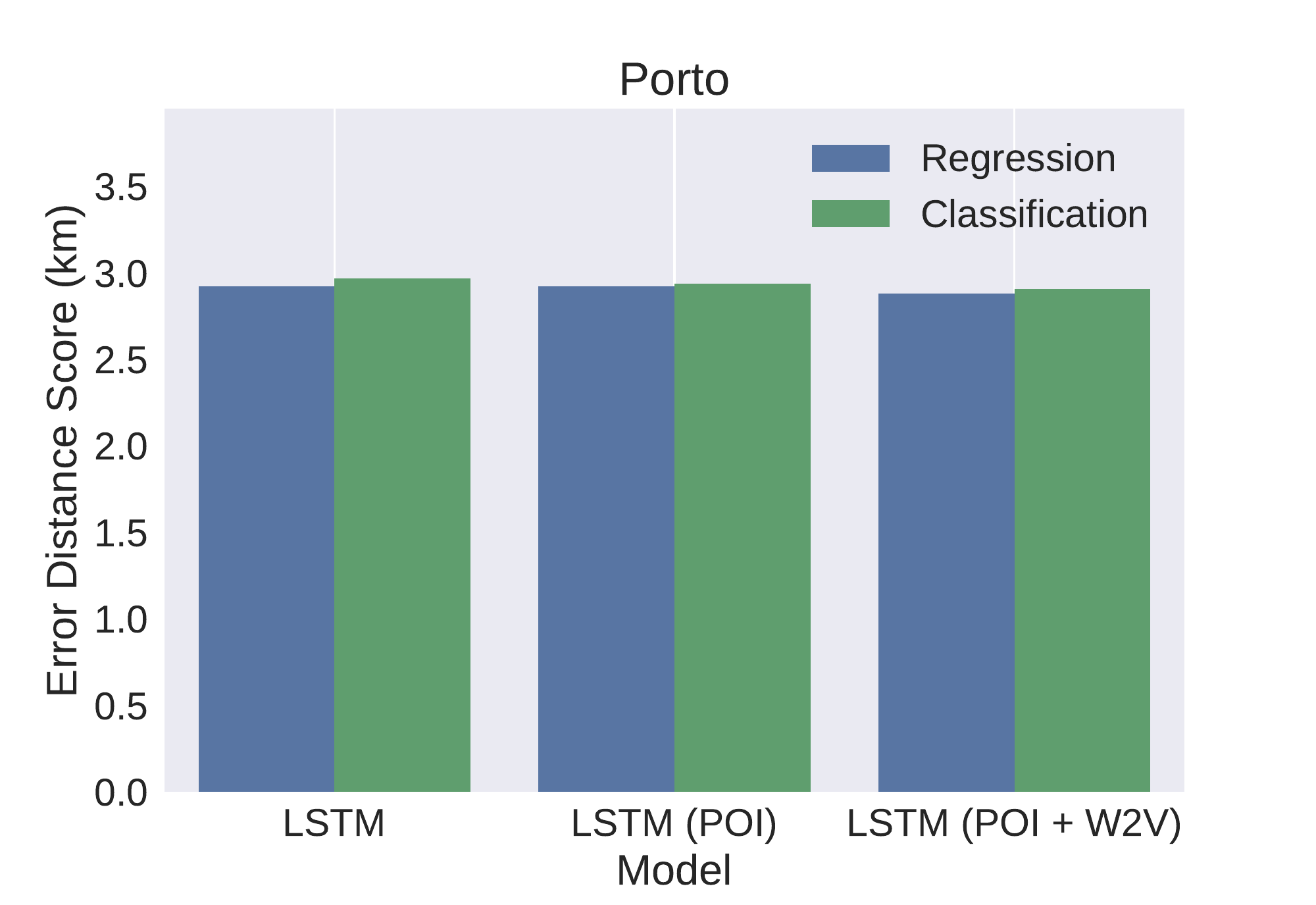}
}
\subfigure{
\includegraphics[width=.31\textwidth]{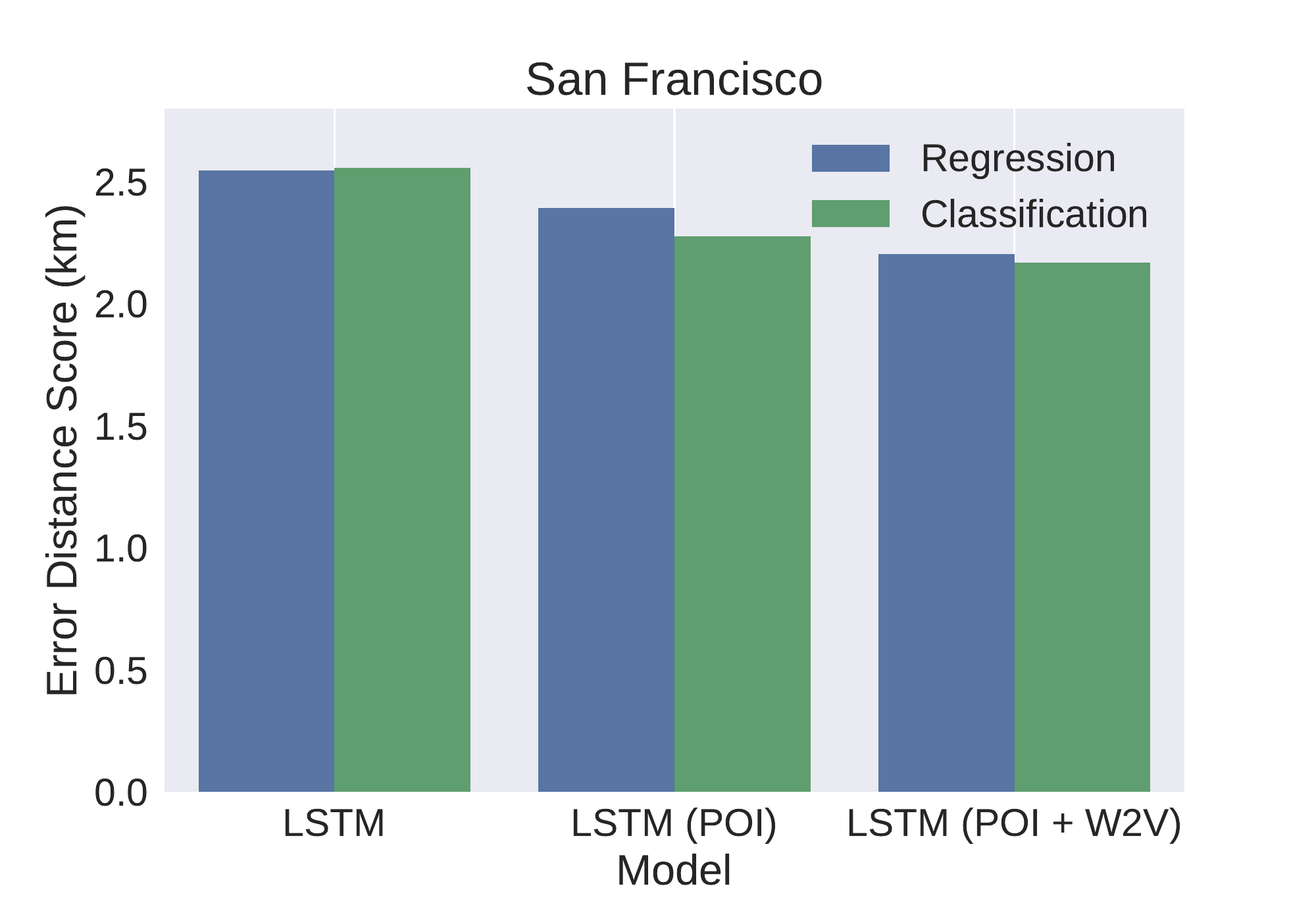}
}
\subfigure{
\includegraphics[width=.31\textwidth]{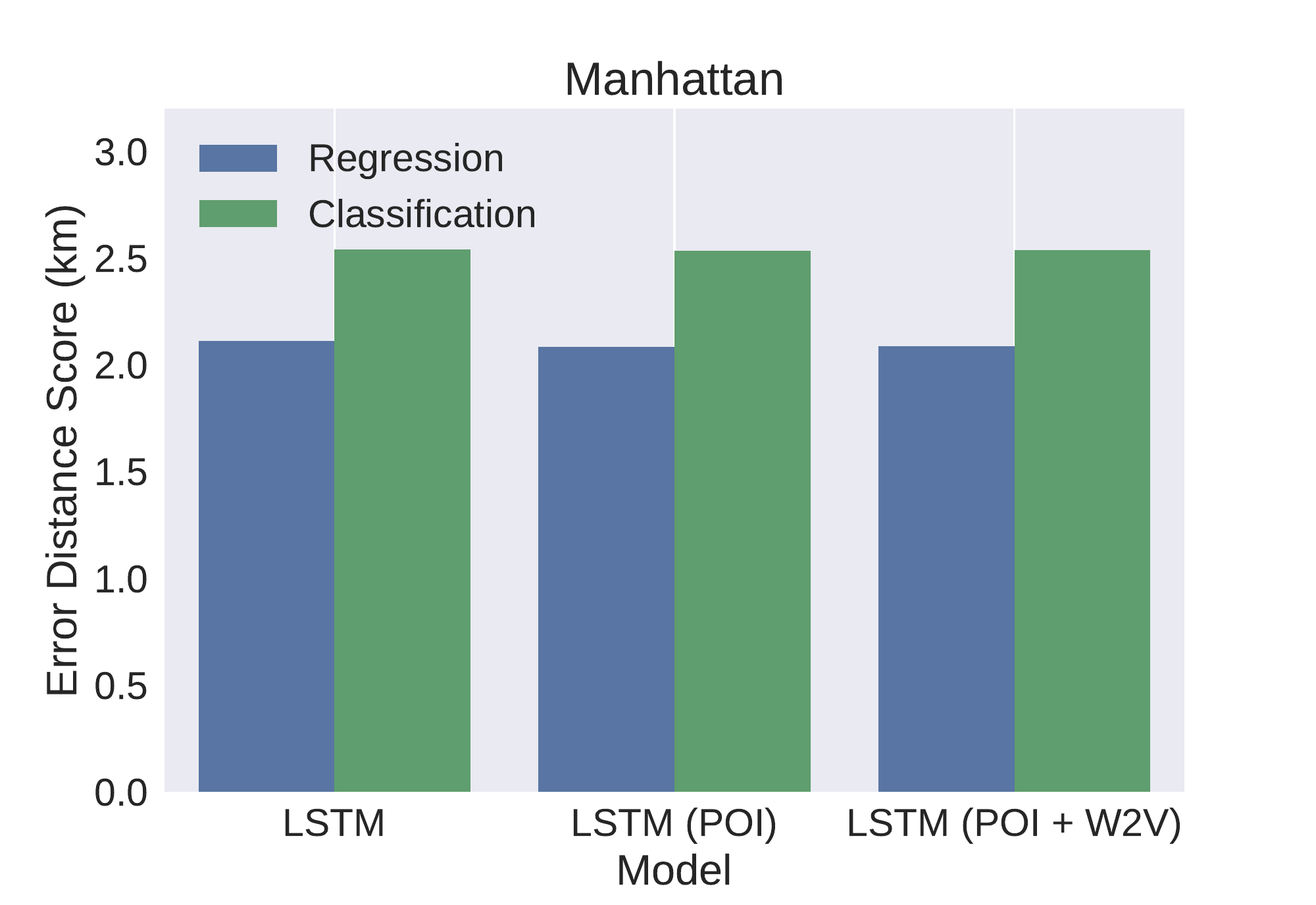}
}
\caption{Classification vs. Regression in Porto, San Francisco and Manhattan.}
\label{fig:regclas}
\end{figure}

All the models reported in Table \ref{result} have been trained on the same set of trajectories. We note that: (i) LSTM (BOC+W2V) reduces the EDS for Porto and Manhattan, outperforming MMLP of 10.5\%, and 18\%, respectively; (ii) in San Francisco, where the trajectories are really dense of coordinates, MMLP obtains better results. However, such a result is due to the oversimplification of the problem, since the model works with the first five and last five trajectory points, meaning that the last input point is really close to the final destination of the taxi. Indeed, MMLP-SEQ, which is the MMLP used with a linearized driver trajectory, is outperformed by 17\% by the LSTM (BOC+W2V); (iii) the usage of spatial semantic features and zone embeddings (BOC+W2V) improves over LSTM; (iv) the results of MMLP in Porto are lower than those obtained in the challenge, due to the small number of trips (only 300) contained in the Kaggle private test set. As explained in the original work \cite{de2015artificial}, such model performs worst than other techniques, e.g, Bidirectional Long-Short-Term-Memory Networks (BLSTM) \cite{thireou2007}, when evaluated on a bigger test set obtained by slicing the training set.

In Figure \ref{fig:regclas} we compare LSTM (BOC+W2V) both in the regression and in the multiclass classification settings. We trained the classification models using the categorical cross-entropy as the loss function and removing the two output neurons. In this way, the output locations are limited to the list of cluster centroids. Finally, the output coordinates are obtained, as in \cite{de2015artificial}, by weighting each cluster centroid through the corresponding probability coming from the softmax layer. The experimental results show that approximating the exact location, i.e., by applying the regression procedure on the latitude/longitude values, allows to reduce EDS in Manhattan, showing the ability of the proposed approach to better perform in cities not equally elongated in latitude and longitude.

\section{Conclusion}
\label{sec7}
In this paper, we have introduced a new way of looking at the problem of the next destination prediction in a taxi journey. Instead of using the complete taxi trajectory, we rely on the individual driver's history, namely the sequence of the last visited points by the driver, i.e., pick-up and drop-off points. Thus, we are able to obtain an instantaneous prediction, just after the starting of the trip, which results in information of paramount relevance for the taxi company dispatcher. This information may improve and optimize fleet management, save costs and reduce waiting times for the customers. In addition, our proposed model captures the taxicab driver's habits, looking at his/her recent history, and this information can be useful for policy makers and urban transportation planners to monitor the impact of taxicab drivers' habits on the whole traffic of a city. 

To the best of our knowledge, we are also the first ones to propose a model that treats this task as a regression problem, instead of a multiclass classification one \cite{yao2017serm,feng2018deepmove}. Finally, we have demonstrated how using (i) geo-located semantic information, such as Foursquare POIs \cite{tong2017simpler}, and (ii) embedding urban zones by considering mobility flows strongly improves the prediction accuracy. 
 
In this perspective, future works will be devoted to integrate different data sources. In particular, the representation of POIs can be enriched by using a greater number of categories, whereas the existing links among POIs, provided by Foursquare, suggest considering POI-graphs instead of isolated points. 
Finally, our approach can be extended to the more general task of the human mobility prediction with respect to both the individual and the collective behaviour, using a deterministic neural network approach instead of a probabilistic model, as in \cite{calabrese2010human}.

\bibliographystyle{splncs} 
\scriptsize{
\bibliography{main}}

\end{document}